\newif\ifarxiv
\arxivtrue 

\ifarxiv
	\documentclass[12pt]{article}
	\usepackage[margin=1in]{geometry}
	\usepackage[slantedGreek]{mathpazo}
	\usepackage[pdftex]{graphicx}
	\usepackage{amsfonts}
	\usepackage{amssymb}
	\usepackage{amsthm}
	\usepackage{graphicx,float}
	\usepackage{amsmath}%
	\usepackage{hyperref}
	\usepackage{natbib}
\else
	\documentclass[12pt]{article}
	\usepackage{enumitem}   
	\usepackage{amsmath,amssymb}
	\usepackage{graphicx}
	\usepackage{caption}
	\usepackage{color}
	\usepackage{dcolumn}
	\usepackage{bm}
	\usepackage{float}
	\usepackage{hyperref} 
	\usepackage{apacite}
	\hypersetup{ colorlinks = true, citecolor = black, urlcolor = blue}

	\makeatletter
	\def\lstlisting@font{\linespread{1}\normalfont\ttfamily}
	\makeatother
	\setenumerate{label=(\alph*),noitemsep, topsep=5pt}
	\definecolor{background-color}{gray}{0.98}

\fi
\DeclareMathOperator*{\argmax}{arg\,max}
\usepackage{listings}
\usepackage{color}
\usepackage{breakcites}

\definecolor{dkgreen}{rgb}{0,0.6,0}
\definecolor{gray}{rgb}{0.5,0.5,0.5}
\definecolor{mauve}{rgb}{0.58,0,0.82}

\title{Deep Learning: Computational Aspects}

\author{Nicholas Polson\thanks{Booth School of Business, University of Chicago}~   and  Vadim Sokolov\thanks{Systems Engineering and Operations Research, George Mason University}}
\date{}

\graphicspath{{./fig/}}
\begin{document}
\maketitle
\ifarxiv
\else
\begin{center}
\subsubsection*{\small Article Type:}
Advanced Review

\hfill \break
\thanks
\fi

\ifarxiv
\begin{abstract}
\else
\subsubsection*{Abstract}
\begin{flushleft}
\fi
In this article we review computational aspects of Deep Learning (DL). Deep learning uses network architectures consisting of hierarchical layers of latent variables to construct predictors for high-dimensional input-output models. Training a deep learning architecture is computationally intensive, and efficient linear algebra libraries is the key for training and inference.  Stochastic gradient descent (SGD) optimization and batch sampling are used to learn from massive data sets. 
\ifarxiv
\end{abstract}
\else
\end{flushleft}
\end{center}
\fi

\ifarxiv
\else
\clearpage
\renewcommand{\baselinestretch}{1.5}
\normalsize
\clearpage
\fi

%
%

\ifarxiv
\section{Introduction}
\else
\section{\sffamily \Large INTRODUCTION} 
\fi

Deep learning (DL) is a form of machine learning that uses hierarchical layers of abstraction to model complex structures. DL requires efficient training strategies and these are at the heart of today's successful applications which range from natural language processing to engineering and financial analysis. While  deep learning has been available for several decades there were only a few practical applications until the early 2010s when the field has changed for several reasons. The renaissance is due to a number of factors, in particular
\begin{enumerate}
	\item Hardware and software for accelerated computing (GPUs and specialized linear algebra libraries)
	\item Increased size of datasets (Massive Data)
	\item Efficient algorithms, such as stochastic gradient descent (SGD).
\end{enumerate}  
The goal of our article is to provide the reader with an overview of computational aspects underlying the algorithms and hardware,  which allow modern deep learning models to be implemented at scale. Many of the leading Internet companies employ DL at scale \cite{hazelwoodapplied}. The most impressive accomplishment of DL is its many successful applications in research and business.  These applications include algorithms such as
\begin{enumerate}
	\item Google Neural Machine Translation \cite{wu2016google} closes the gap with humans in accuracy of the translation by 55-85\% (estimated by people on a 6-point scale). One of the keys to the success of the model is the use of Google's huge dataset.
	\item Chat Bots which predict natural language response have been available for many years. Deep learning networks can significantly improve the performance of chatbots \cite{henderson2017efficient}. Nowadays they provide help systems for companies and home assistants such as Amazon's Alexa and Google home.
	\item Voice Generation was taken to the next level by DL based solutions. Google WaveNet, developed by DeepMind \cite{oord2016wavenet}, generates speech from text and  reduces the gap between the state-of-the art and human-level performance by over 50\% for both US English and Mandarin Chinese.
	\item Google Maps were improved after DL was developed to analyze more than 80 billion Street View images and to extract names of roads and businesses \cite{wojna2017attention}.
	\item Companies like Google Calico and Google Brain Health develop DL for health care diagnostics. Adversarial Auto-encoder model found new molecules to fight cancer.  Identification and generation of new compounds was based on available biochemical data \cite{kadurin2017cornucopia}.
	\item Convolutional Neural Nets (CNNs) have been developed to detect pneumonia from chest X-rays with better accuracy than practicing radiologists \cite{rajpurkar2017chexnet}. Another CNN model is capable of identifying skin cancer from biopsy-labeled test images \cite{esteva2017dermatologist}. \cite{shallue2017identifying} has discovered two new planets using DL and massive data from  NASA’s Kepler Space Telescope.
	\item In more traditional engineering, science applications, such as spatio-temporal and financial analysis DL showed superior performance compared to traditional statistical learning techniques \cite{polson2017deep,dixon2017deep,heaton2017deep,sokolov2017discussion,feng2018deep,feng2018deepa}
\end{enumerate}
In this paper we discuss computationally intensive algorithms for training deep learning models. The main advantage of DL models is the ability to learn complex relationships in high dimensions. However,  a large number of samples is required to make useful predictions. Ability to train DL models using large number (millions) of high-dimensional inputs is the key to the current success of those models. A specialized software that utilizes accelerated hardware computing architectures, such as Graphical Processing Units (GPUs) or Tensor Processing Units (TPU) is typically used to train DL models. We start in Section \ref{sec:dl} by reviewing deep learning models. In Section \ref{sec:opt} we review most commonly used optimization algorithm and highlight that the core operation of those are matrix-matrix multiplications. Is Section \ref{sec:la} we review techniques to accelerate linear algebra operations (e.g. matrix-matrix multiplications). In Section \ref{sec:hard} we review specialized processors that are currently used for training large scale models and the software libraries that support those architectures are discussed in Section \ref{sec:soft}. Finally, we conclude with Section \label{sec:conclusion} where we suggest some further readings. 

\ifarxiv
\section{Deep Learning}
\else
\section{\sffamily \Large Deep Learning}\label{sec:dl}
\fi
Simply put, DL constructs an input-output map. Let $Y$ represent an output (or response) to a task which we aim to solve based on the information in a given  high dimensional input matrix, denoted by $X$. An input-output mapping is denoted by $Y = F(X)$ where $X=(X_1,\ldots,X_p)$ is a vector of predictors. 


\cite{polson2017} view the theoretical roots of DL in Kolmogorov's representation of a multivariate response surface as a superposition of univariate activation functions applied to an affine transformation of the input variable \cite{kolmogorov56,kolmogorov57}. An affine transformation of a vector is a weighted sum of its elements (linear transformation) plus an offset constant (bias). Our Bayesian perspective on DL leads to  new avenues of research including faster stochastic algorithms, hyper-parameter tuning, construction of  good predictors, and model interpretation. 

On the theoretical side,  DL exploits Kolmogorov's ``universal basis''. The fact that DL forms a universal `basis', which we recognize in this formulation, dates back to Poincare and Hilbert.  By construction, deep learning models are very flexible and gradient information can be efficiently calculated for a variety of architectures. On the empirical side, the advances in DL are due to a number of factors, in particular:
\begin{enumerate}
	\item New activation functions, e.g. rectified linear unit ($\text{ReLU}(x) = \max(0,x)$).
	\item Dropout as a variable selection technique and use of multiple layers
	\item Computationally efficient routines to train and evaluate the models as well as accelerated computing on graphics processing unit (GPU) and tensor processing unit (TPU).
	\item Computational software  such as \verb|TensorFlow| or \verb|PyTorch|.
\end{enumerate}


Similar to a classic basis decomposition, the deep approach uses univariate activation functions to decompose a high dimensional $X$. Let $ Z^{(l)} $ denote the $l$th layer, and so $ X = Z^{(0)}$.
The final output $Y$ can be numeric or categorical.
The explicit structure of a deep prediction rule is then
\begin{align}\label{eqn:dl-explicit}
Z^{(1)} & = f^{(1)} \left ( W^{(0)} X + b^{(0)} \right ) \nonumber\\
Z^{(2)} & = f^{(2)} \left ( W^{(1)} Z^{(1)} + b^{(1)} \right ) \\
\ldots  & \nonumber\\
Z^{(L)} & = f^{(L)} \left ( W^{(L-1)} Z^{(L-1)} + b^{(L-1)} \right )\nonumber\\
\hat{Y} (X \mid W,b) & = f^{L+1}(W^{(L)} Z^{(L)} + b^{(L)}) \nonumber\,.
\end{align}
Here $W = (W^{(0)},\dots,W^{(L)})$, are weight matrices and $b = (b^{(0)},\dots,b^{(L)})$ are threshold or activation levels. Designing a good predictor depends crucially on the choice of univariate activation functions $ f^{(l)}$. The $Z^{(l)}$ are hidden features which the algorithm will extract. For a regression problem, we use an identity function for the last layer $f^{L+1} = I$ and for a classification problem, we use a logistic function $f^{L+1}(x) = 1/(1+e^{-x})$.
 For a more extended overview of deep learning models, see~\cite{polson2018deep,lecun2015deep,goodfellow2016deep,schmidhuber2015deep}. From a practical perspective, given a large enough data set of ``test cases",  we can empirically learn an optimal predictor. From a statistical point of view, Equation (\ref{eqn:dl-explicit}) can be viewed as a hierarchical generalized linear model with a  simple GLM being a specific case when $L=0$. When $L>0$, it is practically impossible to interpret neither parameters of the model $(W,b)$ nor the outputs of hidden layers $Z^{(l)}$. Thus, while adding hidden layers allows for learning more complex relations in the data, it prevents us from explaining the prediction rule. Explainability of deep learning models is one of the hurdles that prevents those from being used in heavily regulated industries, such as finance or insurance. It is possible to calculate derivative of the output $\hat Y(X)$ with respect to any of the inputs $X_i$ and thus derive a measure of sensitivity or importance of each of the inputs. The key constraint of this approach is that it is limited to be applied to one input predictor at a time. On the other hand, the main advantage of deep learning model is the ability to capture interactions among the inputs. Thus, the sensitivity analysis is very limiting. Ability to explain predictions of deep learning models is an open area of research. An extension to sensitivity based analysis of deep learning models was proposed by \cite{shrikumar2017learning} who use not only derivatives of the model output but also of the output of each of the hidden layers and derive metrics that allow to explain the interactions among inputs. \cite{sundararajan2017axiomatic} use a modification of the sensitivity approach called integrated gradients to extract explanations of the prediction rules.  \cite{ribeiro2016should} proposed to fit an interpretable model locally around a specific value of the input vector. \cite{ibrahim2019global} propose a method to explain deep learning predictions for different parts of the input space (population of samples). \cite{modarres2018towards} demonstrate empirical performance of several techniques for deep learning model interoperability. 
             
\ifarxiv
\subsection{A Probabilistic View of DL}            
\else
\subsection{\sffamily \large A Probabilistic View of DL}
\fi
Probabilistically the output $Y$ can be viewed as a random variable being generated by a probability  model $p\left(Y\mid  \hat Y(X\mid W,b)\right)$, where $\hat Y(X\mid W,b)$ is a prediction by a deep learning models with weights $W = (W^{(0)},\dots,W^{(L)})$, and $b = (b^{(0)},\dots,b^{(L)})$. Then, given parameters $(W,b)$, the negative log-likelihood defines a loss $\mathcal{L} $ as 
\[
\mathcal{L}(Y, \hat{Y} ) = - \log p\left(Y\mid  \hat Y(X\mid W,b)\right).
\]
Given a training sample $D = \{(X_i,Y_i)\}_{i=1}^n$, the $L_2$-norm, 
\[
\mathcal{L}( Y_i, \hat{Y}( X_i)) = 1/n\sum_{i=1}^n(Y_i - \hat{Y}( X_i))^2_2 
\] 
is traditional least squares, and negative cross-entropy loss is 
\[
	\mathcal{L}( Y_i, \hat{Y}( X_i)) = -\sum_{i=1}^n Y_i \log \hat{Y} ( X_i )
\] for multi-class logistic classification. 

There is a bias-variance trade-off, which is controlled by adding a  regularization term and optimizing the regularized loss
\[
	\mathcal{L}_{\lambda}(Y, \hat{Y} ) = -  \log p\left(Y\mid  \hat Y(X\mid W,b)\right)- \log p( W, b \mid \lambda).
\]

The regularization term is a negative log-prior distribution over parameters, namely
\begin{align*}
- \log p( W, b \mid \lambda) & =  \lambda \phi(W,b),\\
p( W, b \mid \lambda ) & \propto \exp ( - \lambda \phi(W,b)).
\end{align*}
Deep predictors are regularized maximum a posteriori (MAP) estimators, where
\begin{align*}
-p( W, b | D ) & \propto  -p\left(Y\mid  \hat Y(X\mid W,b)\right) p( W, b\mid \lambda) \\
& \propto  \exp \left ( - \log p\left(Y\mid  \hat Y(X\mid W,b)\right) - \log p( W, b) \right ).
\end{align*}
Training requires the solution of a highly nonlinear optimization problem
\[
	\argmax_{W,b} \; \log p( W, b \mid D).
\]
This problem is solved using Stochastic Gradient Descent (SGD) which iteratively updates the parameters $(W,b)$ by taking a step in the direction negative to the gradient. 
The key property is that $ \nabla_{W,b} \log p( W, b \mid D) $ is computationally inexpensive to evaluate using back-propagation algorithms that implemented using modern matrix computation libraries for various hardware architectures. It makes a fast implementation on large datasets possible. {\tt TensorFlow} and {\tt TPUs} provide a state-of-the-art framework for a plethora of deep learning architectures. From a statistical perspective, one caveat is that the posterior is highly multi-modal and providing good hyper-parameter (e.g. number of layers and neurons per layer) tuning can be expensive. This is clearly a fruitful area of research for state-of-the-art stochastic Bayesian MCMC algorithms to provide more efficient algorithms. For shallow architectures,  the alternating direction method of multipliers (ADMM) provides an efficient optimization solution. For more details on probabilistic and Bayesian perspective on deep learning, see~\cite{polson2017}.

\ifarxiv
\section{Optimization Algorithms}\label{sec:opt}
\else
\section{\sffamily \Large Optimization Algorithms}\label{sec:opt}
\fi
We now discuss two types of algorithms for training learning models. First, we discuss stochastic gradient descent, which is a very general algorithm that efficiently works for large scale datasets and has been used for many deep learning applications. Second, we discuss specialized statistical learning algorithms, which are tailored for certain types of traditional statistical models.  

\ifarxiv
\subsection{Stochastic Gradient Descent}
\else
\subsection{Stochastic Gradient Descent}
\fi
Stochastic gradient descent (SGD) is a default gold standard for minimizing
the a function $f(W,b)$ (maximizing the likelihood) to find the deep
learning weights and offsets. SGD simply minimizes the function by
taking a negative step along an estimate $g^k$ of the gradient $\nabla
f(W^k, b^k)$ at iteration $k$. The gradients are available via the
chain rule applied to the superposition of semi-affine functions. 

The
approximate gradient is estimated by calculating
\[
g^k = \frac{1}{|E_k|} \sum_{i \in E_k} \nabla\mathcal{L}( Y_i ,
\hat{Y}( X_i \mid W^{k},b^k)),
\]
where $E_k \subset\{1,\ldots,T \}$ and $|E_k|$ is the number of
elements in $E_k$.\vadjust{\goodbreak}

When $|E_k| >1$ the algorithm is called batch SGD and simply SGD
otherwise. Typically, the subset $E$ is chosen by going cyclically and
picking consecutive elements of $\{1,\ldots,T \}$, $E_{k+1} = [E_k \mod
T]+1$. The direction $g^k$ is calculated using a chain rule (a.k.a.
back-propagation) providing an unbiased estimator of the gradient computed using the entire sample $\nabla f(W^k,
b^k)$. Specifically, this leads to
\[
\mathrm{E}(g^k) = \mathrm{E}\left(\nabla f(W^k, b^k)\right).
\]
At each iteration,  SGD updates the solution
\[
(W,b)^{k+1} = (W,b)^k - t_k g^k.
\]
Deep learning algorithms use a step size $t_k$ (a.k.a learning rate) that is either kept constant or a simple step size reduction strategy, such as $t_k = a\exp(-kt)$  is used. The hyper parameters of reduction schedule  are usually found empirically from numerical experiments and observations of the loss function progression. 

One caveat of SGD is that the descent in $f$ is not guaranteed, or it can be very slow at every iteration. Stochastic Bayesian approaches ought to alleviate these issues. For example,  \cite{wang2019scalable} provide a scalable MCMC algorithm that can be used to train multi-modal loss function that arise when training deep learning architectures.  The variance of the gradient estimate $g^k$ can also be  near zero, as the iterates converge to a solution.  To tackle those problems a coordinate descent (CD) and momentum-based modifications  can be applied. Alternative directions method of multipliers (ADMM) can also provide a natural alternative, and leads to non-linear alternating updates, see \cite{carreira2014distributed}. 

The  CD evaluates a single component $E_k$ of the gradient $\nabla f$ at the current point and then updates the $E_k$th component of the variable vector in the negative gradient direction. The momentum-based versions of SGD, or so-called accelerated algorithms were originally proposed by  \cite{nesterov1983method}. For a more recent discussion, see \cite{nesterov2013introductory}. 
The momentum term adds memory to the search process by combining  new gradient information with the previous search directions. Empirically momentum-based methods have been shown to have better convergence for DL networks \cite{sutskever2013importance}. The gradient only influences changes in the velocity of the update which then updates the variable
\begin{align*}
v^{k+1} =     & \mu v^k - t_k g((W,b)^k)\\
(W,b)^{k+1} = & (W,b)^k +v^k
\end{align*}
The hyper-parameter $\mu$ controls the dumping effect on the rate of update of the variables. The physical analogy is the reduction in kinetic energy that allows to ``slow down" the movements at the minima. This parameter can also be chosen empirically using cross-validation. 

Nesterov's momentum method (a.k.a. Nesterov acceleration) calculates the gradient at the point predicted by the momentum. One can view this as a one-step look-ahead strategy with updating scheme
\begin{align*}
v^{k+1} = & \mu v^k - t_k g((W,b)^k +v^k)\\
(W,b)^{k+1} = & (W,b)^k +v^k.
\end{align*}
Another popular modification \cite{zeiler2012adadelta}, adaptively scales each of the learning parameter at each iteration 
\begin{align*}
c^{k+1} = &      c^k + g((W,b)^k)^2\\
(W,b)^{k+1} = & (W,b)^k - t_k g(W,b)^k)/(\sqrt{c^{k+1}} - a),
\end{align*}
where $a$ is typically a small number, e.g. $a = 10^{-6}$ that prevents dividing by zero. This method is called {\tt AdaGrad}. {\tt PRMSprop} takes the {\tt AdaGrad} idea further and places more weight on recent values of gradient squared to scale the update direction, i.e. we have 
\[
c^{k+1} =  dc^k + (1-d)g((W,b)^k)^2.
\]
The {\tt Adam} method \cite{kingma2014adam} combines both {\tt PRMSprop} and momentum methods and leads to the following update equations
\begin{align*}
v^{k+1} = & \mu v^k - (1-\mu)t_k g((W,b)^k +v^k)\\
c^{k+1} = & dc^k + (1-d)g((W,b)^k)^2\\
(W,b)^{k+1} = & (W,b)^k - t_k v^{k+1}/(\sqrt{c^{k+1}} - a).
\end{align*}
Initial guess in model weights and choice of optimization algorithms parameters plays crucial role in rate of convergence of the SGD and its variants~\cite{sutskever2013importance}.

Second order methods solve the optimization problem by solving a system of nonlinear equations $\nabla f(W,b) = 0$ by applying the Newton's method
\[
(W,b)^+ = (W,b) - \{ \nabla^2f(W,b) \}^{-1}\nabla f(W,b).
\]
Here SGD simply approximates $\nabla^2f(W,b)$ by $1/t$. The advantages of a second order method include much faster convergence rates and insensitivity to the conditioning of the problem. An ill-conditioned problem is the one that has ``flat directions" in which function changes very slowly and it makes SGD rates low. In practice, second order methods are rarely used for deep learning applications \cite{dean2012}. The major disadvantage is their inability to train models using batches of data as SGD does. Second order methods require the inverse Hessian matrix, which in turn requires the entire data set to be calculated.  Since a typical DL model relies on large scale data sets, second order methods become memory and computationally prohibitive at even modest-sized training data sets.

\begin{figure}[H]
	\centering
	\includegraphics[width=0.4\linewidth]{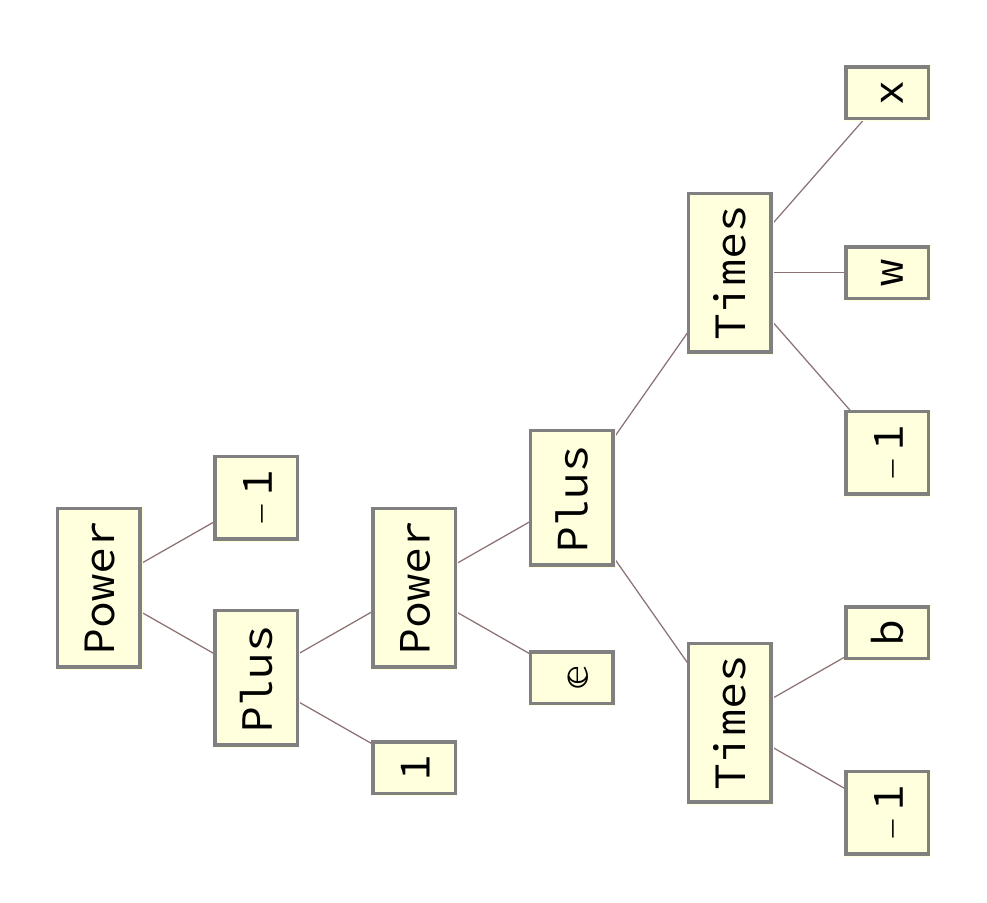}
	\caption{Tree form representation of composition of affine and sigmoid functions: $\frac{1}{e^{-b-wx}+1}$}
	\label{fig:symbolic}
\end{figure}

\ifarxiv
\subsection{Automatic Differentiation (AD)}
\else
\subsection{\sffamily \large Automatic Differentiation (AD)}
\fi
To calculate the value of the gradient vector, at each step of the optimization process, deep learning libraries require calculations of derivatives. In general, there are three different ways to calculate those derivatives. First, is numerical differentiation, when a gradient is approximated by a finite difference $f'(x) = (f(x+h)-f(x))/h$ and requires two function evaluations. However, the numerical differentiation is not backward stable~\cite{griewank2012numerical}, meaning that for a small perturbation in input value $x$, the calculated derivative is not the correct one. Second, is a symbolic differentiation which has been used in symbolic computational frameworks such as \verb|Mathematica| or \verb|Maple| for decades. Symbolic differentiation uses a tree form representation of a function and applies chain rule to the tree to calculate the symbolic derivative of a given function. Figure~\ref{fig:symbolic} shows a tree representation of of composition of affine and sigmoid functions.
The advantage of symbolic calculations is that analytical representation of derivative is available for further analysis. For example, when derivative calculation is in an intermediate step of the analysis. Third way to calculate a derivate is to use automatic differentiation (AD). Similar to symbolic differentiations AD recursively applies the chain rule and  calculates the exact value of derivative and thus avoids the problem of numerical instability. The difference between AD and symbolic differentiation is that AD provides the value of derivative evaluated at a specific point rather than an analytical representation of the derivative. 

AD does not require analytical specification and can be applied to a function defined by a sequence of algebraic manipulations, logical and transient functions applied to input variables and specified in a computer code. AD can differentiate complex functions which involve IF statements and loops, and AD can be implemented using either forward or backward mode. Consider an example  of calculating a derivative of the following function with respect to \verb|x|.
\begin{lstlisting}
def sigmoid(x,b,w):
	v1 = w*x;
	v2 = v1 + b
	v3 = 1/(1+exp(-v2))
\end{lstlisting}

In the forward mode an auxiliary variable, called a dual number, will be added to each line of the code to track the value of the derivative associated with this line. In our example, if we set \verb|x=2, w=3, b=5|, we get the calculations given in Table \ref{tab:AD}.
\begin{table}[H]
	\centering
	\begin{tabular}{l|l}
		Function calculations & Derivative calculations\\\hline
		1. \verb|v1 = w*x = 6| & 1. \verb|dv1 = w = 3| (derivative of \verb|v1| with respect to \verb|x|)\\
		2. \verb|v2 = v1 + b = 11| & 2. \verb|dv2 = dv1 = 3| (derivative of \verb|v2| with respect to \verb|x|)\\
		3. \verb|v3 = 1/(1+exp(-v2)) = 0.99|  & 3. \verb|dv3 = eps2*exp(-v2)/(1+exp(-v2))**2  = 5e-05|\\& (derivative of \verb|v3| with respect to \verb|x|)
	\end{tabular}
	\caption{Forward AD algorithm}
	\label{tab:AD}
\end{table}
Variables \verb|dv1,dv2,dv3| in Table \ref{tab:AD} correspond to partial  (local) derivatives of each intermediate variables \verb|v1,v2,v3| with respect to $x$, and are called dual variables. Tracking for dual variables can either be implemented using source code modification tools that add new code for calculating the dual numbers or via operator overloading.

The reverse AD also applies chain rule recursively but starts from the outer function, as shown in Table~\ref{tab:revad}. 
\begin{table}[H]
	\centering
	\begin{tabular}{l|l}
		Function calculations & Derivative calculations\\\hline
		1. \verb|v1 = w*x = 6| & 4.  \verb|dv1dx =w; dv1 = dv2*dv1dx = 3*1.3e-05=5e-05| \\
		2. \verb|v2 = v1 + b = 11| & 3.  \verb|dv2dv1 =1; dv2 = dv3*dv2dv1 = 1.3e-05| \\
		3. \verb|v3 = 1/(1+exp(-v2)) = 0.99|  & 2. \verb|dv3dv2 = exp(-v2)/(1+exp(-v2))**2;|\\& \verb|   dv3 = dv4*dv3dv2 = 1.3e-05|\\
		4. \verb|v4 = v3| & 1. \verb|dv4=1|
	\end{tabular}
	\caption{Reverse AD algorithm}
	\label{tab:revad}
\end{table}

For DL, derivatives are calculated by applying reverse AD algorithm to a model which is defined as a superposition of functions. A model is defined either using a general purpose language as it is done in \verb|PyTorch| or through a sequence of function calls defined by framework libraries (e.g. in \verb|TensorFlow|). Forward AD algorithms calculate the derivative with respect to a single input variable, but reverse AD produces derivatives with respect to all intermediate variables. For models with a large number of parameters, it is much more computationally feasible to perform the reverse AD.

In the context of neural networks the reverse AD algorithms is called back-propagation and was popularized in AI  by \cite{rumelhart1986learning}. According to \cite{schmidhuber2015deep} the first version of what we call today back-propagation was published in 1970  in a master's thesis \cite{linnainmaa1970representation} and was closely related to the work of \cite{ostrovskii1971uber}. However, similar techniques rooted in Pontryagin's maximization principle \cite{BolGamPon60} were discussed in the context of multi-stage control problems \cite{bryson1961gradient,bryson1969applied}. \cite{dreyfus1962numerical} applies back-propagation to calculate first order derivative of a return function to numerically solve a variational problem. Later \cite{dreyfus1973computational} used back-propagation to derive an efficient algorithm to solve a minimization problem. The first neural network specific version of back-propagation was proposed in \cite{werbos1974beyond} and an efficient back-propagation algo
ritm was discussed in \cite{werbos1982applications}.

Modern deep learning frameworks fully automate the process of finding derivatives using AD algorithms. For example, \verb|PyTorch| relies on \verb|autograd| library which automatically finds gradient using back-propagation algorithm. Here is a small code example using \verb|autograd| library in \verb|Python|.
\begin{lstlisting}
# Thinly wrapped numpy
import autograd.numpy as np
# Basically everything you need
from autograd import grad
# Define a function like normal with Python and Numpy
def tanh(x):
	y = np.exp(-x)
	return (1.0 - y) / (1.0 + y)
# Create a function to compute the gradient
grad_tanh = grad(tanh)
# Evaluate the gradient at x = 1.0
print(grad_tanh(1.0))
\end{lstlisting}

\ifarxiv
\subsection{Architecture Optimization}
\else
\subsection{\sffamily \large Architecture Optimization}
\fi
Currently, there is no automated way to find a good deep learning architecture. An architecture is defined by number of hidden layers, a number of neuron on each layer, parameters that define weight sharing layers, such as convolution layers or recurrent layers. All of those parameters that defined an architecture belong to the set of hyperparameters. Another group of hyperparameters specify the settings for stochastic gradient descent algorithms, e.g. learning rate, momentum, etc.

\begin{figure}[H]
	\centering
	\includegraphics[width=0.35\linewidth]{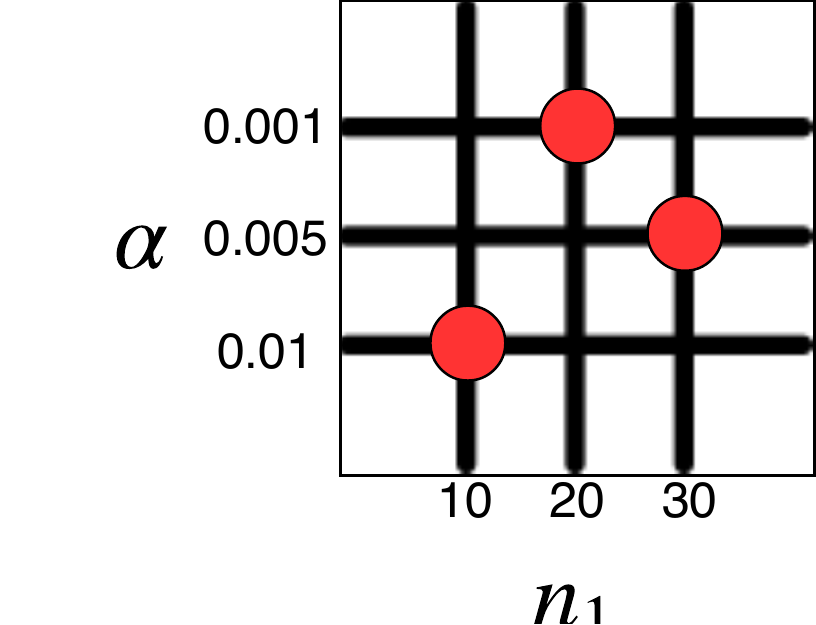}
	\caption{Random grid search for hyperparameters}
	\label{fig:random-search}
\end{figure}

It is not uncommon to use hand-tuning to find a deep learning architecture, when a modeler hand-picks several candidates and choses the one that performs the best on out-of-sample data. It is usually done iteratively and might take weeks or months.  An easiest automated way to find an optimal set of hyperparameters is grid search, when space of hyperparameters is discretized using a grid and a model is estimated for each node of the grid. This approach is used, for example, to find an optimal penalty weight for a LASSO model. However, this approach is not feasible, when number of hyperparameters is large. A random search rather samples from the grid randomly. This, does not guarantee the optimal architecture will be identified but works rather well in practice~\cite{bergstra2012random}. Figure~\ref{fig:random-search} shows an example of randomly chosen grid points, while searching for an optimal  number of neurons on the first hidden layer $n_1$ and the best learning rate $\alpha$.

Bayesian optimization~\cite{srinivas2009gaussian,snoek2012practical} for hyperparameters search is more sample efficient, i.e. requires less model evaluations to find the best candidate. Bayesian methods rely on approximating the relations between hyper-parameters and model performance using  a Gaussian Process surrogate model \cite{mockus2012bayesian}. Gaussian process surrogates have the attractive property that the posterior distribution over function value at any point follows a Gaussian distribution. The stochastic nature of the surrogate allows to quantify uncertainty over the function values and to explore the hyper-parameter space using approaches that can alternate exploration (searching input regions associated with high uncertainty levels of output values) and exploitations (searching regions of local minimal). However, sequential nature of the search process prevents from distributed parallel evaluations of models and is usually less preferred compared to random search when a large number of compute nodes is  available. One can run several instances of Bayesian search in parallel using different initial values \cite{shah2015parallel}. For example, Google's default architecture search algorithm~\cite{golovin2017google} uses batched Bayesian optimization with Mat{\'e}rn kernel Gaussian process. However, this approach, empirically is less efficient compared to random search. Techniques to speed up Bayesian search include early stopping~\cite{gyorgy2011efficient} and using a fraction of the data to evaluate models~\cite{sabharwal2016selecting}.

Genetic-like algorithms provide advantage of sample efficiency and of parallel computing. Recently, \cite{jaderberg2017population} proposed a population based training approach, that evaluates multiple models in parallel and then generates new model candidates by modifying architectures of the models that performed best thus far.

\ifarxiv
\section{Scalable Linear Algebra}\label{sec:la}
\else
\section{\sffamily \Large Scalable Linear Algebra}\label{sec:la}
\fi
The key computational routine required to evaluate a DL model specified by Equation (\ref{eqn:dl-explicit}) is matrix-matrix multiplication. In the context of DL models weights $W$ and inputs and outputs of each layer $X,Z^{(1)},\dots,Z^{(L)},\hat Y$ are called tensors. For example, in image processing input $X$ is a three dimensional tensor, which is made up by three matrices that correspond to red, green and blue color channels. Thus, one of the key operations while training DL or calculating a prediction is a matrix-matrix multiplication, with matrix-vector, dot product or saxpy $a x + y$  (scalar $a$ times $x$ plus $y$) being a special cases.
%
%

Naive implementation of matrix-matrix multiplication would invoke a loop over the elements of the input matrices, which is inefficient. We can parallelize the operations, even on a single processor. Concurrency arises from performing the same operations on different pieces of data is is performed using Single Instruction Multiple Data (SIMD) instructions. SIMD performs multiple independent algebraic operations in one clock cycle. It is achieved by dividing each algebraic operation into multiple simpler ones  with separate hardware in the processor for each of the simple operations. The calculations are performed in a pipeline fashion, a.k.a conveyor belt.  For example, an addition operation can have the following components
\begin{enumerate}
	\item Find the location of inputs
	\item Copy data to register
	\item Align the exponents; the addition \verb|.3e-1+.6e-2| becomes \verb|.3e-1+.06e-1|
	\item Execute the addition of the mantissas
	\item Normalize the result.
\end{enumerate}

When performed one at a time as in a loop, each addition takes 5 cycles. However, when pipelined, we can do it in 1 cycle. Modern processes might have more than 20 components for addition or multiplication operations~\cite{eijkhout2014introduction}.  GPU computing takes it further by using a set of threads for the same instruction (on different data elements), NVIDIA calls it SIMT (Single Instruction Multiple Threads).

Vectorized operations that rely on SIMD or SIMT replace naive loop implementation for calculating the matrix-matrix multiplication.  A vector processor comes with a repertoire of vector instructions, such as vector add, vector multiply, vector scale, dot product, and saxpy. These operations take place in vector registers with input and output handled by vector load and vector store  instructions. For example, vectorization can speedup vector dot product calculations by two orders of magnitude, as shown in code:
\begin{lstlisting}
N = int(1e6)
a = np.random.rand(N)
b = np.random.rand(N)

tic = time.time()
c = np.dot(a,b)
toc = time.time()

print("Vec dot time: " +  str((toc-tic)*1000) + " ms")

c = 0
tic = time.time()
for i in range(N):
	    c+=a[i]*b[i]
toc = time.time()
print("Loop dot time: " +  str((toc-tic)*1000) + " ms")

Vec dot time:  	 1.372 ms
Loop dot time: 435.639 ms
\end{lstlisting}

When calculations are performed on vectors of different dimensions, modern numerical libraries, such as Python's  \verb|numpy| perform those using broadcasting. It involves ``broadcasting'' the smaller array across the larger array so that they have the same shape. The vectorized operation is performed on the broadcasted and the other vector. Broadcasting does not  make copies of data and usually leads to efficient algorithm implementations. For example, to perform \verb|b+z|, where \verb|b| is a scalar and \verb|z| is an  $n$-vector, the numeric library will create a vector of length $n$ \verb|b_broadcast = (b,b,...,b)| and then will compute \verb|(b,b,...,b) + z|.

Further, matrix operations implemented by a linear algebra library take into account the memory hierarchy and aim at maximizing the use of the fastest cache memory which is co-located with the processor on the same board~\cite{eijkhout2014introduction}. In summary, a modeler should avoid loops in their model implementations and always look for ways to vectorize the code. 

Another way a modern DL framework speed up calculations is by using quantization~\cite{tfquant18}, which simply replaces floating point calculations with 8-bit integers calculations. Quantization allows to train larger models (less memory is required to store the model) and faster model evaluations, since cache can be used more efficiently. In some case you'll have a dedicated hardware that can accelerate 8-bit calculations too. Quantization also allows to evaluate large scale models on embedded and mobile devices, and enables what is called edge computing, when data is analyzed locally instead of being shipped to a remote server. Edge computing is essential for Internet of Things (IoT) and robotics systems

\ifarxiv
\section{Hardware Architectures}\label{sec:hard}
\else
\section{\sffamily \Large Hardware Architectures}\label{sec:hard}
\fi
Usage of efficient hardware architectures is an important ingredient in today's success of DL models. Design and optimization of DL hardware systems is currently an active area of research in industry and academia. We currently see an ``arms race'' among large companies such as Google and Nvidia and small startups to produce the most economically and energy efficient deep learning systems. 
\paragraph{GPU Computing}
In the last 20 years, the video gaming industry drove forward huge advances in Graphical Processing Unit (GPU), which is a special purpose processor for calculations required for graphics processing. Since operations required for graphics heavily rely on linear algebra, and GPUs have become widely used for non-graphics processing, specifically for training deep learning models. GPUs rely on data parallelism, when the body of a loop is executed for all elements in a vector:
\begin{lstlisting}
for i in range(10000):
   a[i] = 2*b[i]
\end{lstlisting}
Our data is divided among multiple processing units available, and each processor executes the same statement \verb|a = 2*b| on its local data in parallel. In graphics processing usually the same operation is independently applied to each pixel of an image, thus GPUs are strongly based on data parallelism. The major drawback of GPU computing is the requirement to copy data from CPU to GPU memory which incurs a long latency. Throughput computing, processing large amounts of data at high rates, plays a central role in GPU architectures. High throughput is enabled by a large number of threads and ability to switch fast between them. Modern GPUs would typically have several thousand cores, compare it to the latest Intel i9-family processors that have up to 18 cores. Further, most recent GPUs from NVIDIA would include up to a thousand of so-called tensor cores, that can perform multiply-accumulate operation on a small matrix in one clock cycle. 

Development of GPU code requires skills and knowledge typically not available to modelers. Fortunately, most deep learning modelers do not need to program GPUs directly and use software libraries that have implementations of the most widely used operations, such as matrix-matrix multiplications. 

Currently, Nvidia dominates the market for GPUs, with the next closest competitor being  AMD. Recently, AMD announced the release of a platform called ROCm to provide more support for deep learning. The status of ROCm for major deep learning libraries such as PyTorch, TensorFlow, MxNet, and CNTK is still under development.  

Let us demonstrate the speed up provided by using GPU using a code example:

\begin{lstlisting}
dtype = torch.FloatTensor
N = 50000
x = torch.randn(N,N).type(dtype)
y = torch.randn(N,N).type(dtype)

start = time.time()
z = x*y
end = time.time()
print("CPU Time:",(end - start)*1000)

if torch.cuda.is_available():
    start = time.time()
    x = x.cuda()
    y = y.cuda()
    end = time.time()
    print("Copy to GPU Time:",(end - start)*1000)
    	
    start = time.time()
    a = x*y
    end = time.time()
    print("GPU Time:",(end - start)*1000)
    
    start = time.time()
    a = a.cpu()
    end = time.time()
    print("Copy from GPU Time:",(end - start)*1000)

CPU Time:           11.6
Copy to GPU Time:   28.9
GPU Time:           0.24
Copy from GPU Time: 33.2
\end{lstlisting}

The matrix multiplication operation itself is performed 48 times faster on GPU (11.6 ms vs 0.24 ms). However, copying data from main memory to GPU memory  and back adds another 62.1 ms (28.9 + 33.2). Thus, to efficiently use GPU architectures, it is necessary to minimize amount of data transferred between main and GPU memories 

\paragraph{Intel Xeon Phi}
Recently, in response to the dominance of GPU processors in scientific computing and machine learning, Intel has released a co-processor Intel Xeon Phi. As a GPU,  Xeon Phi provides a large number of cores and has a considerable latency in starting up. The main difference is that Xeon Phi has general purpose cores, while a set of GPU instructions is limited. An ordinary \verb|C| code can be executed on a Xeon Phi processor. However, the ease of use of GPU libraries for linear algebra operations make those the default architecture choice.

\paragraph{DL Specific Architectures}
Companies such as Google or Facebook use deep learning models at extreme scales. Recent computational demand for training and deploying deep learning models at those scales fueled development of custom hardware architectures for deep learning. 

The Intel's Nervana NNP team is focusing on developing a co-processor with fast and reliable bi-directional data transfer. They use a proprietary numeric format called Flexpoint, to increase the throughput. Further, the power consumption is reduced by shrinking circuit size.

Google's Tensor Processing Units (TPU)~\cite{sato2017depth} has two processors, each having 128x128 matrix multiply units (MXU). Each MXU can perform multiple matrix operations in one clock cycle.  Google uses TPUs for all of its online services such as Search, Street View, Google Photos, and Google Translate. TPU uses Complex Instruction Set Computer (CISC) design style which focuses on implementing instructions for high-level complex tasks such as matrix-matrix multiplication with in one clock cycle. In contrast, a typical general purpose CPU follows a Reduced Instruction Set Computer (RISC) design and implements a large number of small primitive instructions (load, multiply,...) and assumes every operation can be represented as a combination of those simple primitives.

There are several other established and startup companies working on developing custom hardware architectures for deep learning computing. Most approaches rely on usage of Field-programmable gate array (FPGA) designs~\cite{brown2012field}. For example, Microsoft's Brainwave~\cite{brainwave17} hardware, which used FPGA is claimed to address the inflexibility of other computing platforms by providing a design that scales across range of data types. Other processor's inflexibility comes from the fact that a set of specific instructions is available at any given architecture.

\ifarxiv
\section{Software Frameworks}\label{sec:soft}
\else
\section{\sffamily \Large Software Frameworks}\label{sec:soft}
\fi
Python is by far the most commonly used language for DL. There are a number of deep learning libraries available, with almost every major tech company backing a different library. Widely used deep learning libraries include \verb|TensorFlow| (Google), \verb|PyTorch| and \verb|Caffe2| (Facebook), \verb|MxNet| (Amazon), \verb|CNTK| (Microsoft). All of those frameworks have \verb|Python| support. For \verb|R| users \verb|Keras| library (\url{https://keras.rstudio.com}) provides a high-level interface to \verb|TensorFlow|, and is the most robust option at this point.

One of the major differences between different  libraries is the use of dynamic vs. static graph computations. Some libraries, such as \verb|MxNet| and \verb|TensorFlow|, allow for both. In static setting a model is fully specified before the training process. In dynamic graphs, structure is defined ``in-thr-fly'' as code gets executed, which is the way our traditional programs are executed. Static graphs provide the opportunity to pre-process the model and to optimize the computations and thus are preferred for large scale models used in production. Dynamic settings provide more flexibility and is typically used during the research and development phase of the model development. Furthermore, dynamic models are easier to debug and easier to learn for those who are familiar with traditional object-oriented programming.

\paragraph{PyTorch} 
PyTorch is native Python library rather than a binding to library written in another language. Thus, it provides an intuitive and friendly interface for Python users to build and train deep learning models on CPU and GPU hardware.  Pytorch is widely used for research as it provides a way to build models dynamically using native Python functions.  On a flexibility-code simplicity scale Pytorch is an attractive option for a researcher who is using Python.

\paragraph{TensorFlow}
TensorFlow is an open source framework written in \verb|C++| with interfaces available for many other languages such as Python.  Although TensorFlow assumes a steeper learning curve when compared to other DL frameworks, its performance on large scale problems across different hardware architectures and support for many popular machine learning algorithms made it a popular choice among practitioners.

\begin{figure}[h]
	\includegraphics[width=1\linewidth]{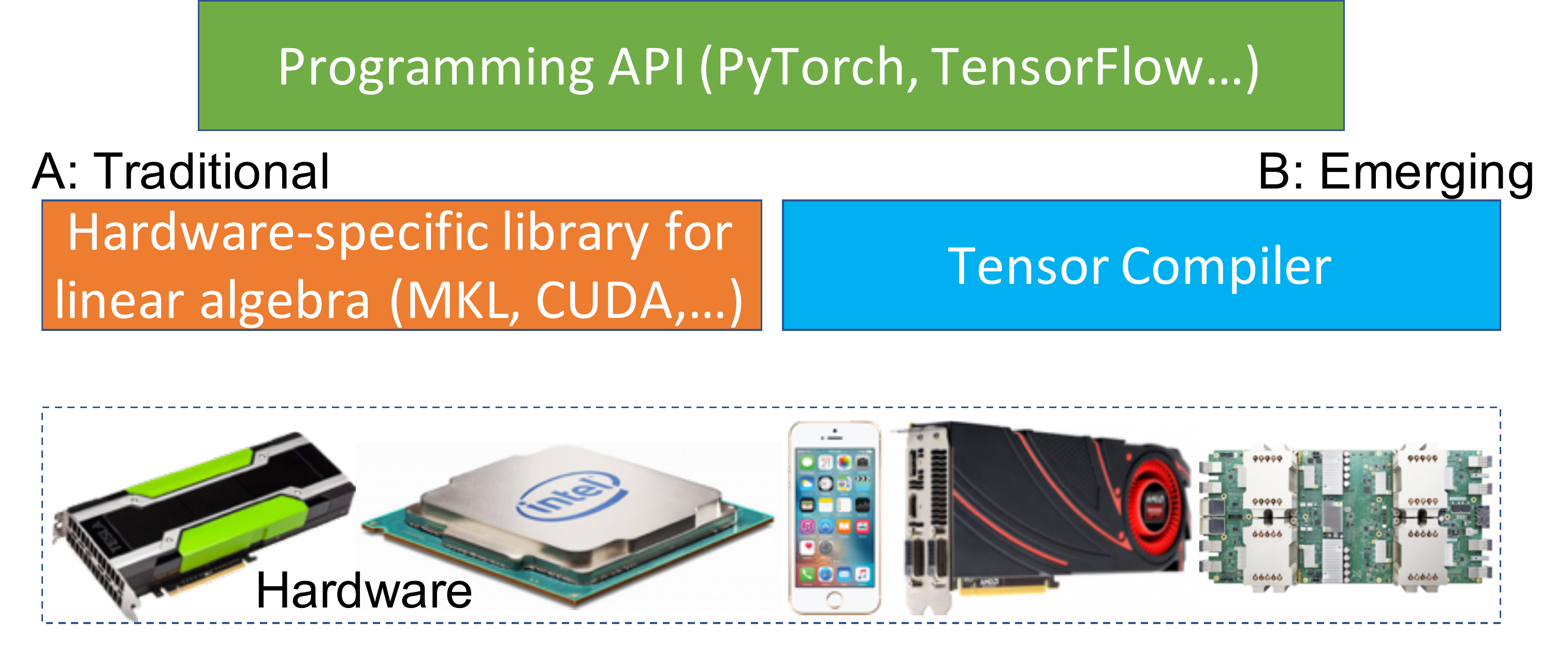}
	\caption{Deep Learning System Hierarchy}
	\label{fig:dlsys}
\end{figure}

\ifarxiv
\subsection{Compiler Based Approach}
\else
\subsection{\sffamily \large Compiler Based Approach}
\fi
Traditional DL systems consists of high-level interface libraries, such as PyTorch or TensorFlow which perform computationally intensive operations by calling functions from libraries optimized for a specific hardware as shown in Figure~\ref{fig:dlsys}. Currently, hardware manufactures have to develop a software stack (a set of libraries) specific to their processors. Nvidia developed CUDA libraries, Intel has MKL library, Google developed TPU library. The reason why Nvidia and not AMD is the GPU of choice for deep learning models is because of Nvidia's greater level of software support for linear algebra and other DL specific computations. 

However, usage of vendor-developed libraries can be limiting. Some expressions might require a complex combination of function calls or might be impossible to write using the functions provided by the vendor library. For example, vendor library might not support sparse matrices. A different approach has recently emerged that relies on compiling linear algebra expressions written in special  language to a code which is optimized for a given hardware architecture. This approach solves two problems, it allows to perform computations that are not implemented in hardware specific-library, and facilitates support for wider a range of architectures, including mobile ones. Recent examples include tensor comprehensions by Facebook~\cite{vasilache2018tensor}, TVM from U Washington~\cite{tvm18}, TACO~\cite{taco17} and Google's XLA (Accelerated Linear Algebra) compiler.

Code below demonstrates impact of \verb|Numba|, which compiles functions written directly in Python. \verb|Numba| uses annotations to compile Python code to native machine instructions. When original python code is mostly performing linear algebra operations, the resulting native machine instructions will lead to performance similar  to \verb|C|, \verb|C++| and \verb|Fortran|. \verb|Numba| generates optimized machine code using the \verb|LLVM| compiler infrastructure. \verb|Numba| supports compilation of \verb|Python| to run on either CPU or GPU hardware and is designed to integrate with the Python scientific software libraries.

\begin{lstlisting}
from numba import jit, double
import math
import numpy as np
import time

@jit(nopython = True)
def mydot(a,b,c):
    for i in range(N):
        c+=a[i]*b[i]
N = int(1e6); a = np.random.rand(N); b = np.random.rand(N)

c = 0
tic = time.time()
mydot(a,b,c)
toc = time.time()
print("Numba dot time: " +  str((toc-tic)*1000) + " ms")

c = 0
tic = time.time()
for i in range(N):
c+=a[i]*b[i]
toc = time.time()
print("Loop dot time: " +  str((toc-tic)*1000) + " ms")

Numba dot time: 	138.628959656 ms
Loop dot time:  	630.362033844 ms
\end{lstlisting}

\ifarxiv
\section{Concluding Remark}\label{sec:conclusion}
\else
\section{\sffamily \Large CONCLUDING REMARKS}\label{sec:conclusion}
\fi
The goal of our paper is to provide  an overview of computational aspects of DL. To do this, we have discussed the core linear algebra, computational routines required for training, and inference using the DL models as well as the importance of hardware architectures for efficient model training. A brief introduction SGD optimization and its variants, that are typically used to find parameters (weights and biases) of a deep learning model is also provided. For further reading, see~\cite{bottou2018optimization}.

Although, DL models have been almost exclusively used for problems of image analysis and natural language processing, more traditional data sets, which arise in finance, science and engineering, such as spatial~\cite{polson2017deep,dixon2017deep} and temporal~\cite{polson2018deepenergy} data can be efficiently analyzed using deep learning. There are a number of areas of future research for Statisticians. In particular, uncertainty quantification and model selection such as architecture design. To algorithmic improvements and Bayesian deep learning.  We hope this review will make DL models accessible for statisticians.

\bibliography{ref}
\end{document}